\documentclass[letterpaper]{article} 
\usepackage{aaai25}  
\usepackage{times}  
\usepackage{helvet}  
\usepackage{courier}  
\usepackage[hyphens]{url}  
\usepackage{graphicx} 
\usepackage{natbib}  
\usepackage{caption} 
\usepackage{booktabs}
\usepackage{amsmath}
\usepackage{amssymb}

\title{To Measure or Not: A Cost-Sensitive, Selective Measuring Environment for Agricultural Management Decisions with Reinforcement Learning}
\author{
    Hilmy Baja, 
    Michiel Kallenberg, 
    Ioannis N. Athanasiadis 
}
\affiliations{
    Artificial Intelligence Group, Wageningen University and Research\\

    hilmy.baja@wur.nl, michiel.kallenberg@wur.nl, ioannis.athanasiadis@wur.nl
}

\begin{document}

\maketitle

\begin{abstract}
Farmers rely on in-field observations to make well-informed crop management decisions to maximize profit and minimize adverse environmental impact.
However, obtaining real-world crop state measurements is labor-intensive, time-consuming and expensive.
In most cases, it is not feasible to gather crop state measurements before every decision moment.
Moreover, in previous research pertaining to farm management optimization, these observations are often assumed to be readily available without any cost, which is unrealistic.
Hence, enabling optimization without the need to have \textit{temporally complete} crop state observations is important.
An approach to that problem is to include measuring as part of decision making.
As a solution, we apply reinforcement learning (RL) to recommend opportune moments to simultaneously measure crop features and apply nitrogen fertilizer.
With realistic considerations, we design an RL environment with explicit crop feature measuring costs.
While balancing costs, we find that an RL agent, trained with recurrent PPO, discovers adaptive measuring policies that follow critical crop development stages, with results aligned by what domain experts would consider a sensible approach. 
Our results highlight the importance of measuring when crop feature measurements are not readily available.
\end{abstract}

%
\begin{links}
    \link{Code}{https://github.com/WUR-AI/CropGym-ToMeasureOrNot}
\end{links}

\section{Introduction}

As the global population continues to grow and the impacts of climate change become more prominent, optimal farm management decisions play a crucial role in our future sustenance. 
Effective farm management practices are key to not only ensure food security but also mitigate the apparent environmental risks related to agriculture  \cite{lipper2014climate}.  
Environmental risks come from mismanagement of farming activities, such as over-applying fertilizer and pesticides \cite{Martinez2021}. 
In practice, farmers rely on their experience and in-field observations to make better-informed management decisions in order to minimize costs and reduce environmental impacts. 
For instance, before fertilizing, a farmer might measure the soil (or leaf) nitrogen content and based on that decide when and how much fertilizer to apply \cite{BERGHUIJS2024127099}.

However, acquiring a large amount of in-field agricultural data is labor intensive, time-consuming and expensive \cite{10.1093/nsr/nwac290}. 
In most cases, the cost and inconvenience of collecting and processing the data outweigh its informational usefulness for the farmer \cite{THOMPSON_Perceptions_2019}.
Furthermore, this issue is compounded in regions with low data availability and high data collection costs, as the infrastructure required for gathering such data is limited, presenting a large hurdle for applying data-driven optimization \cite{cravero2022challenges}.
By integrating data-collection recommendations as part of decision-making, data collection can be optimized and executed only when it is the most beneficial for the farmer.
Specifically, it is of importance to have a system that simultaneously recommends optimal moments in a growing season to obtain measurements of crop states for the objective of cost-effective and optimal management.

The unrealistic assumption that feature observations are readily available has prompted recent work in reinforcement learning (RL) pertaining to active feature measuring for balancing measurement costs \cite{bellinger2021active}. 
Previous work has shown early success in this setting, however, it assumed uniformity in the feature costs \cite{yin2020reinforcement}. 
In the real-world, recognizing that certain features are more expensive to measure than others is essential, as it potentially influences the policies that an RL agent learns.  

In this work, we experiment on the effect of cost in a \textit{measure-and-control} paradigm and design an RL environment in which an agent interacts with a crop growth model (CGM) environment during a growing season.
It is important to acknowledge that crop features have different measuring costs. 
The ability to optimize fertilization with fewer crop state measurements is important to lower the hurdle of applying data-driven methods for improving agricultural management activities.
We summarize our contributions below:
\begin{itemize}
    \item We propose an RL paradigm of costly measurements for crop management, incorporating crop state measurements as part of decision making;
    \item We design and provide code for an RL environment, coupled with the WOFOST \cite{BERGHUIJS2024127099} crop growth model, that allows an agent to simultaneously learn a control policy (fertilization, irrigation, etc.) and a measuring policy. The RL environment is highly configurable, enabling the user to define their own objectives and RL spaces;
    \item We defined a financial focused reward function that includes the definition of feature measurement costs; 
    \item We test our approach \textit{in silico} in a case study in the Netherlands, optimizing fertilization for winter wheat in a rain-fed environment. Experimental results show that costs indeed affect the RL agent's optimization capabilities, where higher costs prove detrimental to the agent's performance. Nevertheless, with realistic costs, the RL agent manages to achieve better performance compared to a baseline of standard practice.
\end{itemize}

\section{Related Work}

This work fits in two growing RL research areas: RL for agriculture and costly active-measuring and -feature acquisition in RL.

\textbf{RL for crop management}. Generally, early RL research in agriculture utilizes simulated environments \cite{GAUTRON2022107182, goldenits2024current}. 
The bulk of prior research builds a Gym interface that wraps around a CGM and obtain management policies \textit{in silico}.
Notable examples include CropGym \cite{cropgym, cropgym2}, gym-DSSAT \cite{gautron2022gymdssat}, SWATGym \cite{madondo2023swat}, and CyclesGym \cite{turchetta2022learning}. 
Related to partially observable crop management, \citet{tao2022optimizing} employed imitation learning to let an RL agent trained with large number of state features to work with a subset of these state features. 
Each work demonstrated a good potential for RL in crop management.

However, these prior works assume \textit{temporally complete feature observations}, which is almost always not the case due to the cost, time and effort of acquiring these observations. 
\citet{turchetta2022learning} identified these costly observations as a core challenge for RL in agriculture. 

\textbf{Costly measuring in RL.} Using deep RL, there have been several studies of costly measurements \cite{bellinger2023dynamic, krale2023act}. 
In the context of active measuring, \citet{nam2021reinforcement} formalized an MDP framework called action-contingent noiselessly observable MDP (ACNO-MDP) that, similarly, defines an explicit cost to observe the complete observation space. 
These prior works assume \textit{completely missing feature observations} in certain time steps, which does not perfectly fit the formalization of our problem, but nonetheless still describes the general joint measuring and control approach.

\textbf{Costly active feature acquisition in RL.} \textit{Active feature acquisition} (AFA) describes a paradigm where an agent can select features to acquire to improve model accuracy. 
There has been a fair amount of prior work in the machine learning domain about AFA \cite{shim2018joint, kossen2023active}. 
In RL, the work of \citet{yin2020reinforcement} includes the formalization of a POMDP extension they call AFA-POMDPs. 
They employ a sequential variational auto encoder (seq-VAE), that is pre-trained with fully observable features offline, to learn a simultaneous sequential measuring and control policy. 
However, they did not investigate the realistic cost of individual features. 
They evaluate the medical sepsis RL environment \citep[from][]{DBLP:journals/corr/abs-1905-05824}, where they assume the cost of acquiring different features to be the same, despite the monetary cost of measuring heart rate and glucose being vastly different. 
This is a critical aspect considering the premise of cost-sensitive feature acquisition. 

To our knowledge, there has been no prior work for costly feature acquisition in RL for crop management. 

\section{Problem Setup}
In this section, we formalize the setting of this work. Moreover, we elaborate in detail the challenge in crop management and the technical implementation of the RL environment.

\subsection{MDP Setting}
\label{MDP}
Crop management problems involve sequential decision-making, which can be formalized into a Markov Decision Process (MDP). 
A standard MDP can be described with the tuple $\mathcal{M} = \langle \mathcal{S, A, T, R}, \gamma \rangle$, where $\mathcal{S}$ is the state space and $\mathcal{A}$ is the action space. 
$\mathcal{T}$ and $\mathcal{R}$ are the environment's transition function $\mathcal{T}(s_{t+1} | s_t, a_t) $ and reward function $\mathcal{R}(s_t, a_t, s_{t+1})$, respectively. 
In crop management problems, as with many real-world environments, the agent is not privy of the complete environment state. 
This is where most real-world MDPs fall into: partially observable MDPs \citep[POMDPs,][]{cassandra1998survey}. 
A POMDP is formalized into the tuple $\mathcal{M} = \langle \mathcal{S, A, T, R, O}, \gamma \rangle$. 
In addition to the standard MDP elements, $\mathcal{O}$ is introduced as the space of possible observations $o \in \mathcal{O}$, which has an observation function of $O(o_{t} | s_{t}, a_t)$.

Coined by \citet{yin2020reinforcement}, \textit{active feature acquisition POMDPs} (AFA-POMDPs) describe an MDP setting where an agent can selectively acquire features at different time steps. 
It is characterized with the tuple $\mathcal{M} = \langle \mathcal{S, A, T, O, R, C}, \gamma \rangle$. 
AFA-POMDPs extend POMDPs by extending the action space $\mathcal{A} = \mathcal{A}_c \times \mathcal{A}_m$ and adding a cost function $\mathcal{C}$.
The action space consists of the control actions $\mathcal{A}_c$ concatenated with measuring actions $\mathcal{A}_m$, and its Cartesian product shows the possible action outputs. 
An action output is denoted as $\mathbf{a}_c \in \mathcal{A}_c$ and $\mathbf{a}_m \in \mathcal{A}_m$. 
$\mathcal{C}$ refers to the cost of unmasking observations. 
In our setup, we assume that $\mathbf{a}_m$ contains a fixed number of features of size $N_m$.

To allow for cost-sensitive, selective measurements, we adapt the above framework by defining: 
\[\mathcal{C}(\mathbf{a}_m) = \sum_{i=0}^{N_m} c_i a_{m_i}\]
$\mathbf{c}$ is a vector that lists explicit costs for individual features $i$, denoted by $c_i$. 
Despite the action space extension, the transition function is similar to the standard MDP with a slight change $\mathcal{T}(s_{t+1} | s, \mathbf{a}_c)$, because we assume measurement actions do not affect state transitions.
The objective of this setup is as follows:

\[
\max_{\pi_c, \pi_m} \mathbb{E} \left[ \sum_{t=0}^{T} \gamma^t \left( \mathcal{R}(s_t, \mathbf{a}_{c,t}) - \sum_{i=0}^{N_m} c_i a_{m_{i},t} \right) \right]
\]

Here, $T$ is a constant, as we defined a fixed sowing and harvest date. Hence, we set $\gamma$ to 1 and the objective effectively becomes the maximum expected total accumulated reward in a trajectory. The explicit reward function $\mathcal{R}$, including the cost term, is defined in the section RL Environment.

In the context of crop management, the elements of the tuples could be mapped as follows: $\mathcal{S}$ is the whole range of crop and environmental states, some which are hidden to the agent. 
Elements in $\mathcal{O}$ are a subset of $\mathcal{S}$; crop and environmental states that the agent can observe. 
$\mathcal{A}_c$ represents the levels of fertilization and $\mathcal{A}_m$ is the possible measurement actions that directly correspond to $N_m$. 
$\mathcal{T}$ is a simulation step of a CGM, $T$ is the simulation duration, $\mathcal{C}$ is the cost of measuring crop features and $\mathcal{R}$ is the amount of projected yield.

\subsection{Crop Management}
\label{crop_management}
As with many biological problems, crop management is highly complex with many factors that influence the process, outcome and yield of the crop growth. 
Generally, in the agricultural domain, this problem is denoted as $G \times E \times M$, which means that crop processes are influenced by the $G$enotype, $E$nvironment and $M$anagement \cite{martin2014}. 
In this work, we focus on arable (\textit{i.e.}, open-field) rain-fed winter wheat. 
Therefore, we do not have any control of the environmental conditions. 
We assume the soil type is uniform through the whole farm. 
Moreover, we assume the absence of yield-reducing factors such as pests and diseases. 
Next, we opt to focus on one variety of winter wheat typically grown in the Netherlands (and western Europe), so we keep the genotype factor fixed. 
In the rain-fed field, we do not apply irrigation. 
This leaves us only with control over nitrogen (N) fertilization for the management.

N is a yield-limiting factor for crops \cite{Chukalla2020}.
Knowing the crop or soil state before applying N fertilizer could enable timely fertilization actions, avoiding waste of resources and detrimental environmental effect. 
Moreover, in dry years, especially for rain-fed environments, the crop experiences water stress and water takes over as the yield-limiting factor, regardless of the N fertilization \cite{eck1988winter}. 

The RL environment we developed is an interface of the Python Crop Simulation Environment \citep[PCSE,][]{PCSE}, which houses several crop growth models. 
In this work, we utilize the World Food Studies \citep[WOFOST,][]{van1989wofost, de201925} CGM. 
WOFOST is a process-based CGM that simulates crop growth. 
It is capable of simulating nitrogen- and water-limited yield production, including the processes that are calibrated to a certain crop variety and location \cite{BERGHUIJS2024127099}. 
WOFOST has been proven to be a robust and reliable CGM, validated through its integral role in the European MARS crop yield forecasting system \cite{van2019mars}.

\begin{figure*}[th]
    \centering
    \includegraphics[width=0.8\linewidth]{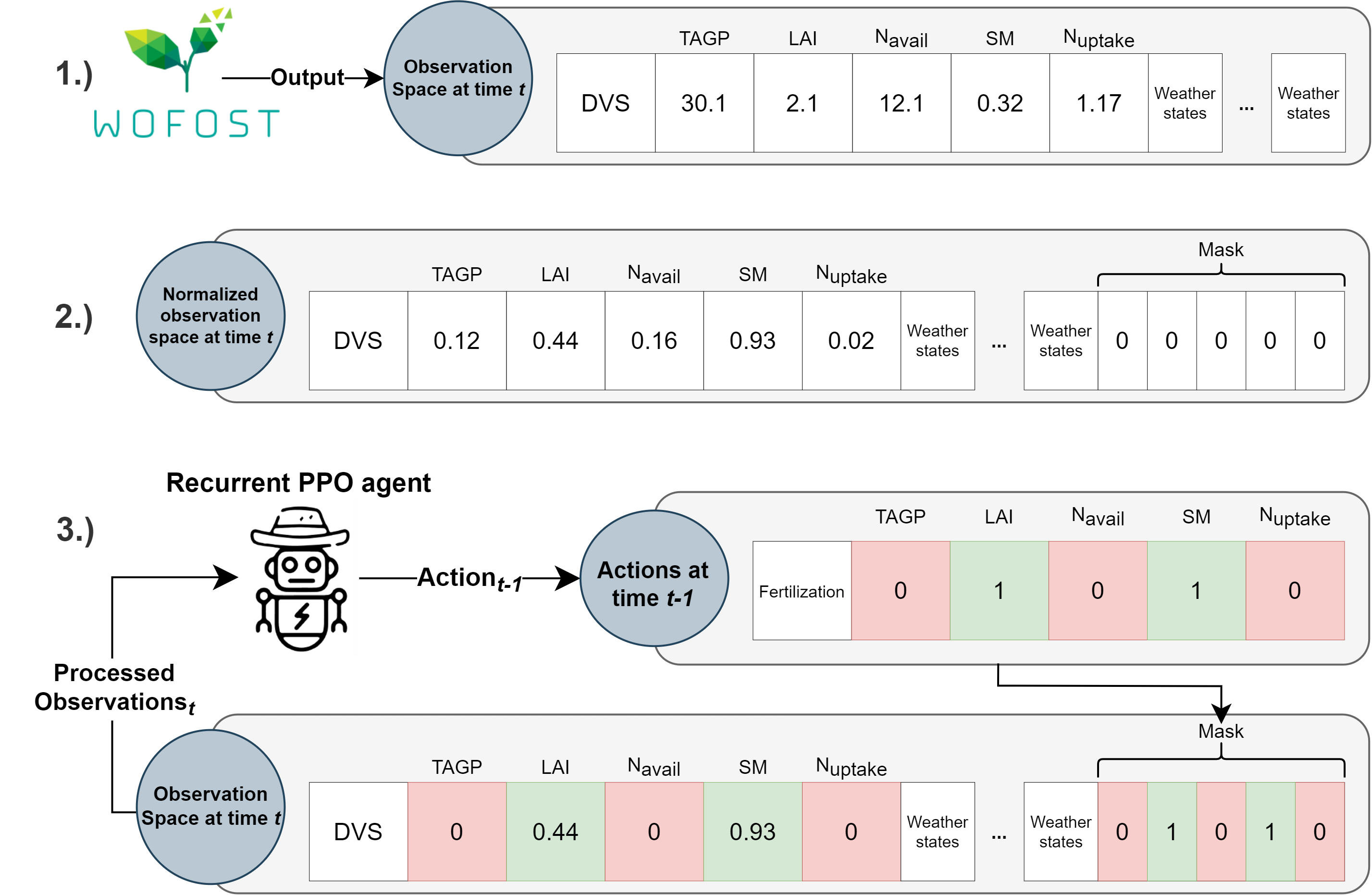}
    \caption{A schematic diagram describing the observation and action space processing, without the \textit{Random} feature.  
    These masks indicate whether the agent decided to measure a specific feature.
    1.) Outputs of WOFOST are processed and flattened into the observation space vector. 2.) The observations are normalized with standardization and observation masks are initialized. 3.) The agent receives a vector comprising of the normalized features and masks. Normalization parameters were derived from multiple random runs. The agent pays a cost for measuring features denoted by the green shade (\textit{LAI} and \textit{SM}). If a feature is not measured, the agent does not pay its cost and the mask along with the normalized feature shows a value of zero.}
    \label{fig:obs}
\end{figure*}

\begin{table*}[ht]

\small
\centering
\begin{tabular}{l l c c c c}
\toprule
Feature         & Description [units] & Fully Observable  & Realistic Cost\\
\midrule
DVS             & Development Stage [-] & $\surd$    &       -   \\
\textbf{TAGP}    & Above Ground Biomass [$kg/ha$] & $\times$ & 25   \\
\textbf{LAI}    & Leaf Area Index [-] & $\times$ & 5  \\
\textbf{NAVAIL}    & Soil Nitrogen Content [$kg/ha$] & $\times$& 20 \\
\textbf{SM}     & Root Zone Soil Moisture [-] & $\times$& 5 \\
\textbf{NuptakeTotal}      & Total Nitrogen Uptake [$kg/ha$] & $\times$ & 20 \\
IRRAD      & Solar Irradiance [$J/m^2/day$] & $\surd$ & - \\
TMIN   & Minimum Temperature [$ ^{\circ} C/day$] & $\surd$ & -  \\
RAIN   & Daily Rainfall [$cm/day$] & $\surd$& -  \\
\midrule
\textbf{Random}      & Random Variable [-] & $\times$ & 10 \\
\bottomrule
\end{tabular}
\caption{Observable crop and weather features and the respective cost to observe. \textit{'Random'} is a distraction feature, which has no correlation to any of the other features. If a feature is fully observable, the agent will receive an observation of it in each time step. We also define the explicit cost for three different experimental setups. One unit of cost is roughly the price of 1 $kg$ of wheat.}
\label{feature-table}
\end{table*}

\subsection{RL Environment}
\label{rl_env}

\subsubsection{RL Spaces:} The observation space dictates the features that the agent can observe and measure. 
Table \ref{feature-table} shows the features of the CGM with continuous values for the crop and weather features. 
Additionally, we add another feature with continuous values called \textit{'Random'} that acts as a distraction feature, which will be explained in the Experiments and Results section. 
Figure \ref{fig:obs} shows a schematic diagram describing the observation and action space processing.

The action space is a vector of discrete and binary values. 
The discrete value corresponds to the agent's control action of applying nitrogen to the crop, which are 7 levels: $\{0, 10, 20, 30, 40, 50, 60\} kg/ha$. 
The binary values correspond to the agent's action of measuring. 
Table \ref{feature-table} shows six features that we define as \textit{'measurable'}. 
The environment has a weekly time step. 
We aggregated the time series data following \citet{cropgym2}: the sequence of weather with a length of $3\times7$ (i.e., daily rain, temperature and solar irradiance) was processed into an average pooling layer, resulting in a vector size of $3\times1$. 
The crop features that had a length of $6\times7$ were shrunk to $6\times1$ by taking its last entry. 
The \textit{Random} feature is also appended to the end of the crop feature at this moment. 
The vectors are then concatenated and flattened, a vector of masks corresponding to the measuring action of the agent at that time step is subsequently appended to the vector, resulting in a vector with a length of $16$ features.

\subsubsection{Constraints:} The nitrogen management regulations in the Netherlands limit the amount nitrogen fertilizer that the farmers are allowed to apply \cite{USDA_2021} due to potential detrimental environmental impact. 
This limit also affects the potential production of the yield. 
Hence, to keep realistic considerations, we impose a constraint for the agent to apply a maximum total of $200 \; kgN/ha$.

\subsubsection{Reward function:} 
The agent's main objective is maximizing yield by balancing the information obtained from costly measurements. 
We designed a natural financial reward function that assumes one unit of reward to be equal to $1 \; kg$ of winter wheat yield. 
$R$ follows:
 \[R_t = ({\mathit{TWSO}_t - \mathit{TWSO}_{t-1})} - \beta N_t - D - \sum_{i=0}^{N_m} c_i a_{m_{i},t}\]
 where $t$ is a weekly time step, $\mathit{TWSO}$ (total weight storage organ) is the wheat yield in $kg/ha$, and $N$ is the amount of Nitrogen in $kg/ha$. 
 The parameter $\beta$ is a ratio of the price of $1 \; kg$ of wheat yield and $1 \; kg$ of $N$ fertilizer, so we set $\beta = 2$ to mimic the respective prices \citep{Agri23b, Agri23a}. 
 $D$ is a deployment cost that penalizes the agent for going out the field to apply $N$, with a penalty of 10 if $N_t > 0$.
 $c_i$ is the cost of measuring a specific feature (\textit{i.e.} \textit{LAI}, \textit{SM}, \textit{etc.}) and $a_{m_i,t}$ denotes a measurement action $a_{m_i}$ at time step $t$. We describe the rationale for the costs in the next section.

\textbf{RL agent.} For this experiment, we employ a PPO agent with recurrent networks (LSTM-PPO), as implemented by the library Stable Baselines 3 \cite{stable-baselines3}.
We take advantage of the recurrent properties of the agent to discover temporal dependencies in the environment \cite{POMDP2021baseline}.
Also, in their experiments, \citet{bellinger2021scientific} proved that masks in the observation space work well for costly measuring with an actor-critic agent.
The agent has two separate LSTM networks for actor and critic, each with 2 hidden layers with size of $256$ and an activation function of $\tanh$. 
We set the agent's learning rate to $1\mathrm{e}{-4}$. The rest of the hyperparameters we kept same as the default.

\section{Design Rationale and Assumptions}
\label{rationale}
\textbf{Feature selection and observation.} WOFOST has a plethora of crop states/features that describe various crop and soil processes. 
The RL agent only observes a subset of these features, which makes the environment inherently partially observable. 
Though, a large portion of the features are highly correlated and redundant. 
Hence, we handpicked $6$ crop features that are considered most important for the task of nitrogen fertilization, out of which, $5$ are \textit{measurable} (Table \ref{feature-table}). 
\textit{TAGP} describes the biomass of the crop, which is highly correlated to yield. \textit{LAI} describes the area of leaves, thus explaining the photosynthesis capabilities of the crop. 
\textit{NuptakeTotal} describes the total amount of N the crop took from the soil. 
\textit{SM} describes the moisture around the crop's root, related to potential water stress for the crop. 
\textit{NAVAIL} describes the amount of N in the soil. 
We give the agent full access to the remaining crop feature: development stage (DVS). 
This is akin to the farmer going out and visually checking a crop if it has emerged from the ground ($DVS \geq 0$), has reached the flowering stage ($DVS \geq 1$) or matured ($DVS = 2$). 
\textit{DVS} grows monotonically from $-0.1$ to $2$, hence it can be used to infer not only the stages of crop growth, but also the progression of time during the crop growth period.
So, we also use \textit{DVS} as a proxy for time and let the agent learn critical crop stages for adaptive measuring and fertilizing strategies.
The agent also has full access to the weekly weather information, as this akin to a farmer checking the weather of last week from their nearest weather station.

\textbf{Feature measurement costs.} We defined explicit cost to obtain measurements for different features. 
We base the cost of an estimation on how expensive it is to obtain a measurement of the feature (shown under Table \ref{feature-table}). 
\textit{SM} and \textit{LAI} are cheaper since it is possible to measure through sensors \cite{hummel2001soil}, various remote sensing \cite{hasegawa2010improving, schmugge1983remote} or optical methods \cite{LAI_10.1093/jxb/erg263}. 
\textit{NuptakeTotal} can be measured with non-destructive sensors \cite{ulissi2011nitrogen}, which requires considerable labour to do for a whole field. 
\textit{NAVAIL} can be measured with soil tests \cite{soil_test}, requiring the soil to be sent to a lab, consequently incurring monetary cost and time. 
\textit{TAGP} is an important feature for estimating yield and growth process of crops \cite{ma2022tagp}. 
The work of \citet{kuyah2015optimal} evaluated methods and the cost to measure above-ground biomass. 
Measuring \textit{TAGP} requires considerable labor, and could be done through destructive or non-destructive methods.

\textbf{Assumptions.} We have made a few assumptions regarding the implementations and experimental design:
\begin{itemize}
    \item \textit{Measurements are noiseless.} Noise is almost always present due to either sensor, human or environmental factors. However, in this work we assume the measurements of the RL agent are noiseless.
    \item \textit{Measurement actions do not alter the underlying model state.} All measurements are assumed to be done non-destructively.
    \item \textit{Measurements are only repeatable costs.} Some measurements can be obtained by paying a \textit{one-time cost}, such as buying a soil sensor for \textit{SM}. However, we assume all measurements are \textit{repeatable costs}, which is realistic due to the labor and time required to process the acquired measurements.
    \item \textit{If measured, the feature values are immediately available.} Some measurements require processing or waiting time, especially for soil tests.
\end{itemize}

\begin{table*}[t]
\centering
\begin{tabular}{lrrrrrr}
\toprule
Scenario & LAI & SM & NuptakeTotal & TAGP & NAVAIL & \textit{Random} \\
\midrule
No-cost &  \textbf{23.9} (4.0) & \textbf{25.2} (5.6) & \textbf{19.9} (4.4) & \textbf{24.2} (4.2) & \textbf{26.9} (4.7) & \textbf{23.5} (4.3) \\
Flat-cost & \textbf{2.2} (1.7) & \textbf{1.8} (1.6) & \textbf{2.2} (1.3) & \textbf{2.0} (1.5) & \textbf{1.5} (1.2) & \textbf{1.0} (0.9) \\
Realistic & \textbf{5.5} (2.5) & \textbf{4.8} (2.5) & \textbf{0.6} (0.9) & \textbf{0.2} (0.3) & \textbf{0.3} (0.4) & \textbf{0.2} (0.3) \\
Exp-cost & \textbf{0.0} (0.0) & \textbf{0.0} (0.0) & \textbf{0.2} (0.3) & \textbf{0.0} (0.0) & \textbf{0.1} (0.1) & \textbf{0.0} (0.0) \\
\bottomrule
\end{tabular}
\caption{Number of times  the agent performed a measuring action for the corresponding feature in a one-year period. We report the average (and mean absolute deviation) across years and seeds. The maximum amount of measuring actions an agent can perform in one year is $47$.}
\label{measure_sum_table}
\end{table*}

\begin{table}[th]
\centering
\begin{tabular}{llc}
\toprule
Scenario & \multicolumn{2}{c}{Yield [$tons/ha$]} \\
 & Median (IQR) & 95\% CI \\
\midrule
No-cost & \textbf{7.86} (0.97)  & (6.92, 9.40) \\
Flat-cost & \textbf{7.59} (1.17) & (6.88, 8.98)\\
Realistic & \textbf{7.46} (1.09)  & (6.45, 9.13) \\
Exp-cost & \textbf{6.63} (4.35)  & (0.93, 8.94) \\
\midrule
All-observed & \textbf{7.86} (0.96)  & (6.99, 9.44) \\
None-observed & \textbf{7.21} (3.82)  & (0.93, 8.81) \\
\midrule
Standard-practice & \textbf{7.30} (1.09)  & (6.55, 8.65) \\
Random-spread & \textbf{0.65} (0.28)  & (0.26, 1.06) \\
\bottomrule
\end{tabular}
\caption{Yield for different cost scenarios across seeds and evaluation years in the Netherlands.}
\label{yield_table}
\end{table}

\section{Experiments and Results}
\label{experimental_setup}

In this section we would like to answer the following questions:
\begin{enumerate}
    \item How does different measurement costs affect a recurrent PPO agent's ability to optimize yield?
    \item How adaptive is the RL agent's measuring policies between cold and hot years?
    \item Does the agent learn to ignore features with no information?
\end{enumerate}

\subsubsection{Training conditions:} We conduct experiments where the recurrent PPO agent has a joint task of measuring and applying N fertilizer in the WOFOST CGM environment. 
Training was done in with semi-fine soil and climate conditions of the Netherlands with $3$ coordinates: $(52, 5.5), (51.5, 5), (52.5, 6)$ $(^\circ N, ^\circ E)$ obtained from the NASA POWER weather dataset \cite{sparks2018nasapower}. 
The agent trained on the odd years from 1990 to 2022 (n=$16$), and we save the even years for evaluation. 
We set the simulation length to be fixed for $47$ weeks, from sowing (October 1st) until harvesting (September 1st). 
We evaluate the agent with even years (n=$16$), and weather from the Netherlands with coordinates $(52.5, 5.5) (^\circ N, ^\circ E)$. 
In total, one evaluation run contains $16$ episodes.

The initial soil conditions (\textit{i.e.}, how much moisture and N nutrients were in the soil when sowing) affects the agent's learned policy. 
An agent will learn that the soil conditions always start with a similar amount, potentially remembering this and discouraging it from measuring. 
Hence, to add per episode variability to the RL environment, for soil moisture and nitrogen content, we set it with values from a randomized normal distribution generator, with a mean and standard deviation of $15$ (in $kg/ha$ for nitrogen content). 
The values are clipped to be bounded between $0$ and $100$.
The generator is seeded, therefore repeatable across different seeds.

\subsubsection{Training scenarios:} To understand the effect of measuring cost in maximizing yield, we train agents with several cost scenarios. 
First, a \textit{Realistic} cost scenario, which was explained in section Design Rationale and Assumptions and the explicit costs shown in Table \ref{feature-table}. 
Second, a \textit{Flat-cost} scenario, where each feature has a cost of 10 ($kg$ of wheat) to measure, a bit lower than the average cost of \textit{Realistic} features. 
Third, a \textit{No-cost} scenario, where the agent is free to perform measure actions. 
And fourth, an \textit{Exp-cost} scenario, a very expensive \textit{flat} cost of 60 for measuring each feature, forcing the agent to converge to a policy of maximizing reward without measuring. 
To compare the performance of the RL agents to a simple fertilizing policy, we add a non-measuring fixed-fertilizing policy that fertilizes 3 times in fixed dates, each two months apart from January to May. Each fertilization has an amount of 66.67 $kg/ha$, with a cumulative total of 200 $kg/ha$ adhering to the constraint applied in our experiments.
Additionally, we add a \textit{random-spread} baseline that randomly spreads 200 $kg/ha$ Nitrogen fertilizer throughout the growing seasons.
We treat \textit{Exp-cost} as a baseline of what an agent can achieve without measuring in an AFA-POMDP setting. 
On the other hand, we treat \textit{No-cost} as the upper bound for attainable yield.
Further, we show two scenarios of RL agents trained outside the AFA-POMDP framework, with only a fertilizing action, without the task of measuring.
Consequently, these scenarios have an easier learning process due to the smaller action and observation space.
\textit{All-observed} was trained with complete features.
\textit{None-observed} was trained with only \textit{DVS} and the weather features, without any crop features.
We train each scenario with $1.5M$ steps and repeat the experiments $10$ times with consistent seeds.

\begin{figure*}[t]
\centering
\includegraphics[width=1.0\textwidth]{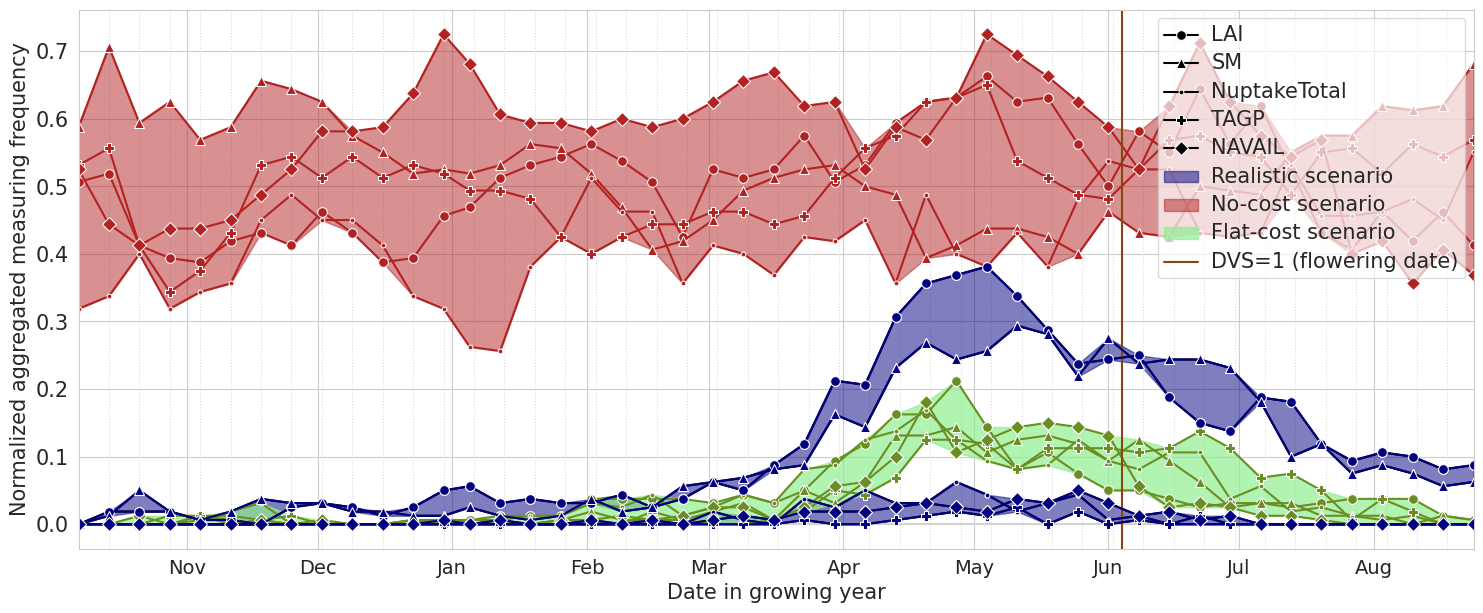}
\caption{The normalized aggregated measuring frequency calculated from all seeds and years. We show shaded bands to emphasize the measuring action frequency throughout the growing year. The different colors represent the cost scenarios, and the different marker shapes represent measured features. The vertical line shows the average flowering day.}
\label{figure-measuring-period}
\end{figure*}

\subsubsection{Attained yield and measuring frequency of each scenario:} In Table \ref{yield_table}, we report the yield across different seeds and evaluation years. 
We estimate the 95\% confidence interval around the sample median by bootstrapping. 
Additionally, we compute the inter-quartile range from the sample to show variance between years.
In Table \ref{measure_sum_table} we report the measuring frequency by showing the average number of measurement actions performed by the agent in a one-year period.
Moreover, Figure \ref{figure-measuring-period} shows the temporal measuring policy of each cost scenario.
We designed a reward function that maximizes yield through N fertilization, though we constrain the amount of N it can apply. 
At first glance to Table \ref{yield_table}, we immediately notice that scenarios with lower cost achieve better yield.
All scenarios learned to apply the maximum allowed amount of N fertilizer, \textit{i.e.,} $200 \; kg/ha$. 
Notwithstanding, each scenario achieved different yields. We provide analyses below:

Between the four cost scenarios, \textit{No-cost} achieves a higher median yield compared to the others and the lowest yield variance. 
The agent typically measures roughly half of the available steps in a year.
Also, its performance is almost identical to \textit{All-observed}, which hints to the potential redundancy for an agent in a crop management environment to have \textit{temporally complete} observations.

There is a notable difference in achieved yield between \textit{Flat-cost} and \textit{Realistic}, with the former being ahead. 
Though, both are lower than \textit{No-cost}, indicating that the presence of measuring cost prohibits the agent to obtain the required state information to perform optimal and timely fertilization. 
Moreover, the result of \textit{Flat-cost} suggests that, with the unrealistic assumption that each features have the same cost to measure, the agent can utilize the more expensive features to better estimate the overall state. 
In \textit{Realistic}, the difference in cost is mirrored in the measurement frequencies of certain features. 

From an agronomic point of view, the measuring policy is understandable; the agent measures total N uptake (\textit{NuptakeTotal}) twice as much as N soil content (\textit{NAVAIL}), despite the similar cost, since it gives a better idea of how much N the crop took from the soil. 

The expensive flat cost scenario (\textit{Exp-cost}) performed the worse and has the largest variance between episodes, highlighting the importance of measuring. 
We note that the \textit{Exp-cost} still converged to measuring some features in rare cases (i.e. \textit{NuptakeTotal} and \textit{NAVAIL}). 
The agent measured features that are related to N fertilization to potentially check whether more N is needed.

The agent trained in scenario \textit{All-observed} achieved virtually identical results as \textit{No-cost}.
Scenario \textit{None-observed} performed worse than \textit{Realistic}, showing that the observations of these crop states are necessary for the agent to make informed decisions for timely fertilization actions.

In scenarios that achieve lower yields, a significant amount of fertilizer is wasted and not converted into yield. 
This excess N application will have a significant environmental impact. 
These results highlight the significance of measuring to achieve more effective and optimal fertilization policies. 

\subsubsection{Random feature measuring policy:} We added \textit{Random} as a distraction feature to investigate how an agent will deal with a feature that carries no information. 
We set the \textit{Realistic} cost to $10$, same as \textit{Flat-cost}. 
A seeded normal distribution generator sets the values of \textit{Random}, with a mean and standard deviation of $10$. 
In the scenarios with costs (in Table \ref{measure_sum_table}), we notice that \textit{Random} is the feature measured the least. 
However, in \textit{No-cost}, it is measured about the same amount compared to other features. 
We notice in the \textit{Realistic} scenario, the \textit{Random} variable is measured as often as biomass (\textit{TAGP}), despite the large difference in cost, suggesting that the agent is able to distinguish features that carry valuable information. 

\subsubsection{Temporal measuring policy:} We present Figure \ref{figure-measuring-period} to show when the agent decides to measure in a growing year. 
The agents can fully observe the crop development stage (DVS), which we use as a proxy for time.
We notice with \textit{No-cost}, there is no clear pattern of measuring, which can be associated with the agent having no incentive to balance measuring costs with fertilization costs. 
In the scenarios with costs, the agent typically starts measuring close to the crop's flowering stage, which is a critical stage for grain (yield) formation. 

\begin{figure}[ht]
\centering\
\includegraphics[width=0.47\textwidth]{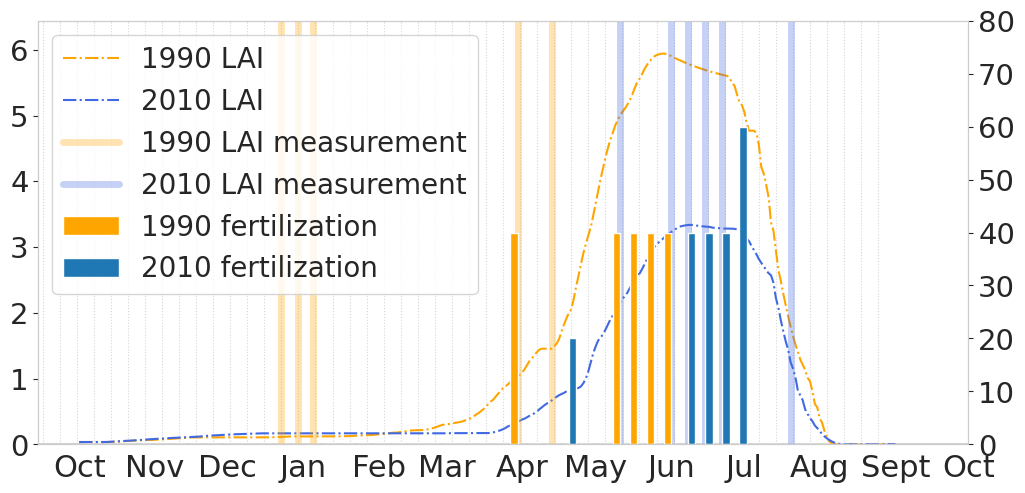}
\caption{\textit{LAI}, measurement and fertilization actions for 1990 and 2010 growing year period for the \textit{Realistic} scenario. The y-axis (left) shows \textit{LAI} development and secondary y-axis (right) shows bars depicting applied fertilizer actions in [$kg/ha$]. Measurement actions are shown with transparent vertical lines.}
\label{figure-adaptive}
\end{figure}

\subsubsection{Adaptive policy:} In this section we show how the agent adapts its fertilization based on its measurements and crop development. 
In colder years, the development of the plant is delayed, so there is a need to adjust management. 
We report Figure \ref{figure-adaptive}, showing how a cold and normal year affects the measuring and fertilization policy in two scenarios.
Years 1990 and 2010 had yearly cumulative minimum temperatures of $1980.11^\circ C$ and $1488.36^\circ C$, respectively. 
Note that in 2010, \textit{LAI} experienced delayed development due to a colder winter. 
Since the agent always observes \textit{DVS} and the weather, it expects delayed crop development and adjust its measuring schedule for colder years. 
Accordingly, after obtaining the crop state information, the agent starts fertilizing later to adapt to the delayed development. 
Note that in 1990, it chose to measure three times in January, which is generally the emergence period for winter wheat, signifying the start of stem and leaf (\textit{LAI}) development. 
This demonstrates that the agent learned a policy that is able to anticipate growth throughout the season.

\section{Discussion and Limitations}


\subsubsection{Realistic cost policy: }The scenarios \textit{No-cost} and \textit{Flat-cost} achieved the best performance as they allow access to relatively complete crop state information \cite{wu2022optimizing}.
However, these cost scenarios are unrealistic; crop state measurements are not always readily available, nor do they have uniform cost.
The \textit{Realistic} scenario agent, which had to pay a realistic cost for measuring features, still managed to perform well. The agent frequently measured \textit{LAI}, which is strongly related to nitrogen leaf content in different crop stages \cite{yin2003leafnitrogen}, and \textit{SM} (soil moisture), which is a major yield-limiting factor \cite{day1970some, fahad2019regional}. 
With these results, we demonstrate the importance of measuring when crop feature data is not readily available or difficult to obtain consistently.

\subsubsection{RL Algorithm: } We employed LSTM-PPO for our problem setup. \citet{yin2020reinforcement} utilize LSTM-A3C to train their RL policy, of which they employ a  seq-VAE that is pre-trained with fully observable features that they feed to the LSTMs hidden states. In contrast, we do not employ pre-training in our approach and let the agent learn from a \textit{tabula rasa}. A performance increase is probable if we pre-train our LSTM-PPO networks with fully observable features. Nevertheless, we show that our simple approach obtains measuring policies that enable good yield.

\subsubsection{Assumptions:} In section Design Rationale and Assumptions we explain the rationale of our design choices and defined some limiting assumptions of our proposed solution.
Future work may build on our findings by relaxing these assumptions such as incorporating temporal delays or adding noise to measurements.

\subsubsection{Scalability: } A limitation of our approach is the need to possibly retrain the RL agent for different sites and setups. Scaling up is possible by including various sites in the agent's training. Though, it is challenging from both domain perspectives: generalization remains a challenge in RL and ML \cite{cobbe2019quantifyinggeneralizationreinforcementlearning, Li2022WhyRG}; and site-specific N management demands detailed site knowledge of in-field N soil variability for accurate recommendations \cite{schut2020soil}. In this work, we evaluate our approach in a representative case study with a well-calibrated CGM, thus we leave further exploration of this challenge for future work.

\subsubsection{Sim2real gap:} The experiments in this work are done \textit{in silico}, hence there exists a gap between simulation and reality.
Various things are not simulated in modern CGMs: yield-reducing factors such as pests and diseases \cite{donatelli2017pest}, genetic variability \cite{hirel2007genetic}, among others, which introduce distribution shifts between simulation and reality.
RL algorithms that are robust to distributional shifts \cite{turchetta2022learning} and randomization \cite{tobin2017random} can help narrow this gap.
Nevertheless, a CGM can achieve high accuracy if calibrated specifically to the year and conditions of the location it is simulating \cite{ahmed2020models, he2017data}. 
For future work we intend to test and evaluate our developed system in field trials.

\section{Conclusion}

In this work, we propose an RL approach of integrating data collection to decision making by obtaining a measuring policy that balances measuring costs with fertilization.
Inspired by the problem of difficult and costly data collection in agriculture, we design an RL environment with realistic considerations by adapting the framework of \textit{AFA-POMDPs}.
We evaluate our approach \textit{in silico} in a case study in the Netherlands, with WOFOST, a thoroughly validated CGM.
Our test includes different cost scenarios.
We find that the RL agent discovers adaptive measuring policies that coincide with critical crop development stages and learns that some features are more valuable than others.
While cost does indeed affect the discovered measuring policy, with realistic costs the agent is still capable of achieving good yield by utilizing cheaper measurements.

Our work highlights the importance of measuring when crop feature measurements are not readily available.
By integrating measurement recommendations in the decision making process, we can minimize unnecessary data collection. 
This approach is an important step to lower the hurdle of applying data-driven optimization for crop management.
Ultimately, our work allows for the ability to optimize yield while reducing required data samples, further realizing better crop management policies for mitigating adverse environmental impact and sustaining global food demands.

\section{Acknowledgments}
The authors thank Ron van Bree for constructive discussions during the design stage of this work, and Aike Potze for the useful comments during the writing stage of the manuscript. This work was supported by the European Union Horizon Research and Innovation programme, Smart Droplets (Project Code: 101070496).

\bibliography{aaai25}

\end{document}